\documentclass{article}

\usepackage{arxiv}
\usepackage{times}
\usepackage{epsfig}
\usepackage{graphicx}
\usepackage{amsmath}
\usepackage{amssymb}
\usepackage{authblk}

\usepackage{color}
\usepackage{subcaption}
\usepackage{booktabs}
\usepackage{comment}
\usepackage{algorithm}
\usepackage{algcompatible}
\usepackage{multirow}
\usepackage{diagbox}
\usepackage{sidecap}
\usepackage{enumitem}
\usepackage[accsupp]{axessibility}

\usepackage[normalem]{ulem}

\usepackage{tabularx}
\newcolumntype{C}[1]{>{\centering\arraybackslash}p{#1}}

\usepackage{amsmath,amsfonts,bm}

\def\eqref#1{equation~\ref{#1}}

\def\1{\bm{1}}

\def\vf{{\bm{f}}}

\def\mE{{\bm{E}}}

\def\mM{{\bm{M}}}

\def\mQ{{\bm{Q}}}
\def\mR{{\bm{R}}}

\def\mU{{\bm{U}}}
\def\mV{{\bm{V}}}

\DeclareMathAlphabet{\mathsfit}{\encodingdefault}{\sfdefault}{m}{sl}
\SetMathAlphabet{\mathsfit}{bold}{\encodingdefault}{\sfdefault}{bx}{n}
\newcommand{\tens}[1]{\bm{\mathsfit{#1}}}

\def\tC{{\tens{C}}}

\def\tE{{\tens{E}}}

\def\tQ{{\tens{Q}}}

\def\tX{{\tens{X}}}

\def\sI{{\mathbb{I}}}

\newcommand{\R}{\mathbb{R}}

\usepackage[pagebackref=true,breaklinks=true,colorlinks,bookmarks=false]{hyperref}
\usepackage[capitalise]{cleveref}

\newcommand{\acro}{\textsc{C-Pic}}
\newcommand{\acroshort}{\textsc{C-Pic}}

\begin{document}
	\title{Cherry-Picking Gradients: Learning Low-Rank Embeddings of Visual Data via Differentiable Cross-Approximation}
	
    \author[1]{Mikhail Usvyatsov}
	\author[1]{Anastasia Makarova}
	\author[2]{Rafael Ballester-Ripoll}
	\author[3]{Maxim Rakhuba}
	\author[1]{Andreas Krause}
	\author[1]{Konrad Schindler}
	\affil[1]{ETH Zürich}
	\affil[2]{IE University}
	\affil[3]{HSE University}
	
	\maketitle
	
	\begin{abstract}
	    We propose an end-to-end trainable framework that processes large-scale visual data tensors by looking \emph{at a fraction of their entries only}.
	    Our method combines a neural network encoder with a \emph{tensor train decomposition} to learn a low-rank latent encoding, coupled with \emph{cross-approximation} (CA) to learn the representation through a subset of the original samples.
		CA is an adaptive sampling algorithm that is native to tensor decompositions and avoids working with the full high-resolution data explicitly. Instead, it actively selects local representative samples that we fetch out-of-core and on demand.
		The required number of samples grows only logarithmically with the size of the input.
		Our implicit representation of the tensor in the network enables processing large grids that could not be otherwise tractable in their uncompressed form.
		The proposed approach is particularly useful for large-scale multidimensional grid data (e.g., 3D tomography), and for tasks that require context over a large receptive field (e.g., predicting the medical condition of entire organs). The code is available at \url{https://github.com/aelphy/c-pic}.
	\end{abstract}
	\vspace{-0.5em}
	\section{Introduction}
	\label{intro}
	
	%
	%

	%
	Over the past decade, convolutional neural networks (CNNs) in combination with parallel processing on GPUs have brought about dramatic improvements in machine learning for image data.
	%
	%
	Unfortunately, parallel hardware is memory-limited, leading to a \emph{curse of dimensionality}: state-of-the-art 2D network architectures are typically not viable for data with 3 or more dimensions, because one runs out of memory to store the corresponding tensors.
	Despite efforts to mitigate the problem via sparse convolutions \cite{choy20194d,3DSemanticSegmentationWithSubmanifoldSparseConvNet,Hackel2020} or octrees \cite{riegler2017octnet,wang2017cnn},
	one must in practice limit the size of the inputs. E.g., the upper bound for 3D volumetric data is about $512^3$ voxels on high-end commodity hardware.
	
	It is well-known that visual data usually lives on lower-dimensional manifolds and, therefore, is in principle compressible; c.f. classical ideas like Eigenfaces~\cite{turk1991} or non-negative tensor factorisation~\cite{shashua2005}. This motivates us to seek a more memory-efficient representation of high-dimensional visual data that is more efficient, and at the same time compatible with the gradient-based learning process of neural networks.
	
	\begin{figure}[!t]
		\centering
		\includegraphics[width=\linewidth]{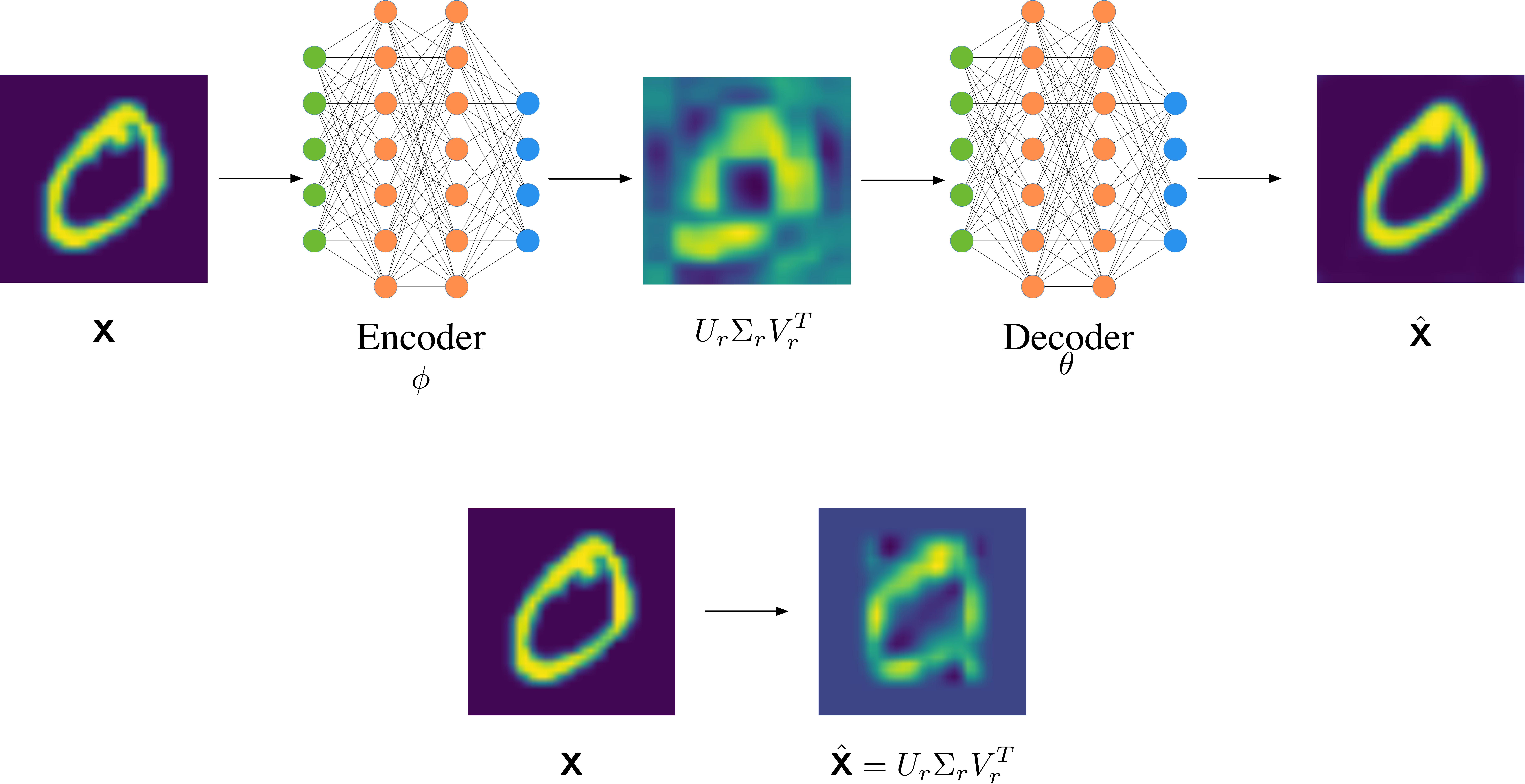}
		\caption{2D illustration of learned low-rank embedding: rank-3 compression of the input with SVD (the matrix equivalent of TT decomposition) severely degrades the image (\emph{bottom row}). In contrast, our encoder warps the image such that the same rank-3 truncation loses little information and can be decoded almost perfectly (\emph{top row}).}
		\label{fig:SVDexample}
	\end{figure}
	
	Selecting a sampling resolution at which data is recorded and/or processed is always a trade-off between resources (memory, run-time, power, etc.) and the level of detail and context that the algorithm has access to.
	Indeed, relatively small tensors are sufficient for applications where either high-frequency details are not crucial and one can operate at low spatial resolution (e.g., face recognition), or long-range context has little impact and one can process local windows (e.g., character recognition).
	But some tasks do require sharp details and long-range context.
	For instance, it has been shown that 3D object classification performance improves with increasing resolution~\cite{Hackel2020}.
	A similar situation arises when making holistic predictions from medical imagery: high-resolution detail helps to better spot subtle tissue changes, whereas the global context is needed to assess the extent of the condition.
	The ever-increasing resolution of the scanning hardware will only exacerbate this discrepancy -- even current CT or MRI scanners, with typically $1024\times1024\times128$ voxels, are at the limit of what can be conveniently processed.
	
	In this paper, we propose \acro\ (for "cherry-picking gradients"), a framework for learning with tensors while looking only at a small fraction of their entries.
	\acro \ exploits the fact that, after a suitable non-linear mapping, the learned representation can be constrained to have low rank. The constraint gives rise to a smart sampling strategy that adaptively selects which tensor entries to be shown to the architecture. The whole pipeline is end-to-end trainable with back-propagation, so that the learned, low-dimensional embedding is optimally tuned to a given prediction task.
	Crucially, our approach can operate out-of-core, meaning that it does not need to store the full input tensors in memory, but only small (hyper-)cubes around the sampled locations.
	It can therefore handle massive spatial resolutions that are orders of magnitude larger than the available memory, particularly on GPUs (we have experimented with volumes up to $8192^3 \approx 0.5 \cdot 10^{12}$ voxels).

	The key insight underlying our novel representation is related to non-linear dimensionality reduction:
	if we can transform the tensor values in a way that "flattens the manifold",
	then we can explicitly impose a \emph{low rank} structure on the representation, which we do with the tensor train (TT) decomposition~\cite{oseledets2011tensor}.
	I.e., we learn an end-to-end function that maps the input data to the desired output via a low-rank TT bottleneck.
	This is possible due to two important properties of the TT decomposition:
	\emph{(i)} one can perform basic tensor arithmetic in the compressed format, as well as back-propagate through the decomposition;
	and \emph{(ii)} there exist efficient \emph{cross-approximation (CA)} algorithms that find an approximate TT decomposition based only a small set of samples,
	rather than the complete input~\cite{oseledets2010tt}.
	While there have been attempts to use the TT format within a neural network~\cite{novikov2015tensorizing}, our work is, to the best of our knowledge, the first to employ cross-approximation for learning; making it possible to operate at high spatial resolution without running into memory limits.

With classical manifold learning, our work shares the assumption of an underlying low-dimensional, but non-linear data manifold. However, our embedding is discriminative, in the sense that the projection onto the manifold is learned end-to-end, taking into account the prediction task. In this way, the learned encoder minimises not the error when reconstructing from the latent representation, but the error of the desired output after decoding the representation.
	See ~\cref{fig:SVDexample} for a 2D illustration.
	By keeping the input and the activation maps of the encoder implicit, we circumvent what is arguably the main limitation of grid representations of dimension $D\geq 3$: their huge memory consumption, exponential in $D$.
	To summarise, our {\bf contributions} are:
	\begin{enumerate}[leftmargin=15pt,topsep=4pt]
		\setlength{\itemsep}{0pt}
		\setlength{\parskip}{2pt}
		\item We design a first end-to-end neural architecture for high-dimensional, but low-rank visual data that exploits tensor decompositions;
		\item We develop a computational scheme for back-propagating through cross-approximation. The differentiable CA step allows one to learns an optimal embedding from a limited number of sample evaluations and thereby opens the door to very large resolutions.
		\item We develop an iterative basis projection scheme to project the learned TT features onto a canonical basis, so that they can serve as a basis for regression tasks.
	\end{enumerate}
	
	We demonstrate our approach on two different medical image analysis problems and show that we perform on par with the state-of-the-art. Furthermore, \acro\ with the same hyper-parameters can work on double the resolution while other state-of-the-art methods fail due to the memory limit. 
	
	\section{Background and Related Work}
	\label{rel_work}

	\subsection{Tensor Train Decomposition}
	
	Tensors are a fundamental data structure for computer vision in the current age of deep learning. For our purposes, a tensor $\displaystyle \tX \in
	\R^{I_1 \times \dots \times I_D}$ is a discrete sampling of a $D$-dimensional space on a grid $\mathbb{I} = I_1 \times \dots \times I_D$, with $I_d$ samples along dimension $d$.
	
	For a long time, low-rank approximations of matrices have been used in computer vision as a tool to compress, classify, or restore visual data, e.g.~\cite{andrews1976,sirovich1987,turk1991,zhang2013,indyk2019}.
	However, they build on matrix-specific decomposition techniques like SVD or independent component analysis, which do not directly generalise to tensors of dimension \textgreater2. 
	
	
	More recently, low-rank priors have been generalized to the tensor case; see also~\cref{app:tensor_decompositions}. The model used in this paper, the tensor train (TT)~\cite{oseledets2011tensor}, decomposes a tensor of dimension $D$ into $D$ 3-dimensional tensors. Consequently, its number of parameters grows only {\em linearly} with the dimensionality.
	The TT is defined as
	\begin{equation}
		\label{tt_idx}
		\displaystyle
		\tX[i_1, \dots, i_D] = 
		\tQ_1[1, i_1, :]
		\tQ_2[:, i_2, :]
		\dots
		\tQ_D[:, i_D, 1],
	\end{equation}
	where the tensors $\displaystyle \{\tQ_d\}_{d=1}^D, 
	\tQ_d \in \R^{r_{d-1} \times I_d \times r_d}$, are called \textit{TT-cores} and $r_d$ are the \emph{TT-ranks} ($r_0\!=\!r_D\!=\!1$).
	The TT decomposition has $\mathcal{O}\big(D\cdot(\max_d[r_d])^2\cdot\max_d[I_d]\big)$ storage cost. Importantly, basic linear algebra operations such as linear combination of tensors can be carried out directly in this format without prior decompression (i.e., recomposing the cores).
	
	Robust numerical schemes exist to find the TT decomposition. The standard TT-SVD algorithm yields a quasi-optimal decomposition~\cite{oseledets2011tensor} but is based on multiple rounds of singular value decomposition(SVD), i.e., it must visit all entries of the input tensor.
	Of crucial importance for our work is a different algorithm, known as \emph{cross-approximation}, that efficiently constructs the TT-cores based on an adaptively chosen sequence of local samples from the input tensor. Only a small fraction of all tensor elements need to be queried; see Section~\ref{sec:method}.
	
	\subsection{Applications in Machine Learning}
	
	Tensor decompositions have been investigated as a way of extracting features from high-dimensional datasets~\cite{phan2010tensor,bengua2015optimal}, and at large scale~\cite{fonal2019distributed}. The Tucker decomposition, in particular, has recently also been extended to nonlinear interactions between the cores, with either Gaussian Processes \cite{Zhe2016} or deep neural networks
	\cite{liu2017deepcp}. 
	
	\cite{ballester2019tthresh} explore the Tucker decomposition as a lossy compression tool for multi-dimensional grid data.
	Our work goes further: we share the aim to compress gridded data via the low-rank representation, but learn an encoder/decoder structure tailored to the rank-constrained bottleneck to minimise the associated information loss.
	
	In deep learning, the TT format has so far been used mostly to compress very large network layers \cite{novikov2015tensorizing}.
	Recently, the format was employed as part of a conditional generative model for drug design \cite{kuznetsov2018,zhavoronkov2019deep}, where a variational auto-encoder was combined with a TT-induced prior over the joint distribution of latent variables and class labels. There, a global set of TT-cores are learnable parameters, while we TT-decompose each individual input tensor, thus requiring an efficient and differentiable procedure.
	
	%
	
	%
	
	\subsection{Prediction of Health Indicators}
	
	In section \cref{sec:experiments}, we demonstrate our approach on the concrete target application of predicting a patient's future condition from medical 3D scans (CT of the lung and MRI of the brain, respectively).
	Regressing health indicators from scan data has a long tradition in medical image analysis, e.g.,~\cite{jerebko2003multiple, yang2019development}. Following the general trend in computer vision, recent methods mostly employ deep CNNs for the task.
	Examples include brain age estimation from MRI scans, e.g.,~\cite{huang2017,bermudez2019}; and survival prediction from MRI scans, e.g.,~\cite{isensee2017brain, feng2020brain}.
	All these works use standard CNN architectures like VGG, U-Net or ResNet, and operate on low-resolution scans (sizes below 200$\times$200$\times$100 voxels) to stay within GPU memory limits.
	
	\section{Method}
	\label{sec:method}
	
	We first describe our model in feed-forward mode, where it maps tensor-valued input data to the prediction via a low-rank TT bottleneck.
	Then, we explain the efficient implementation and end-to-end learning of this model, including back-propagation through the cross-approximation algorithm, and a projection of TT-cores to obtain a unique feature representation. 

	\subsection{Model Architecture}
	
\begin{figure*}[t!]
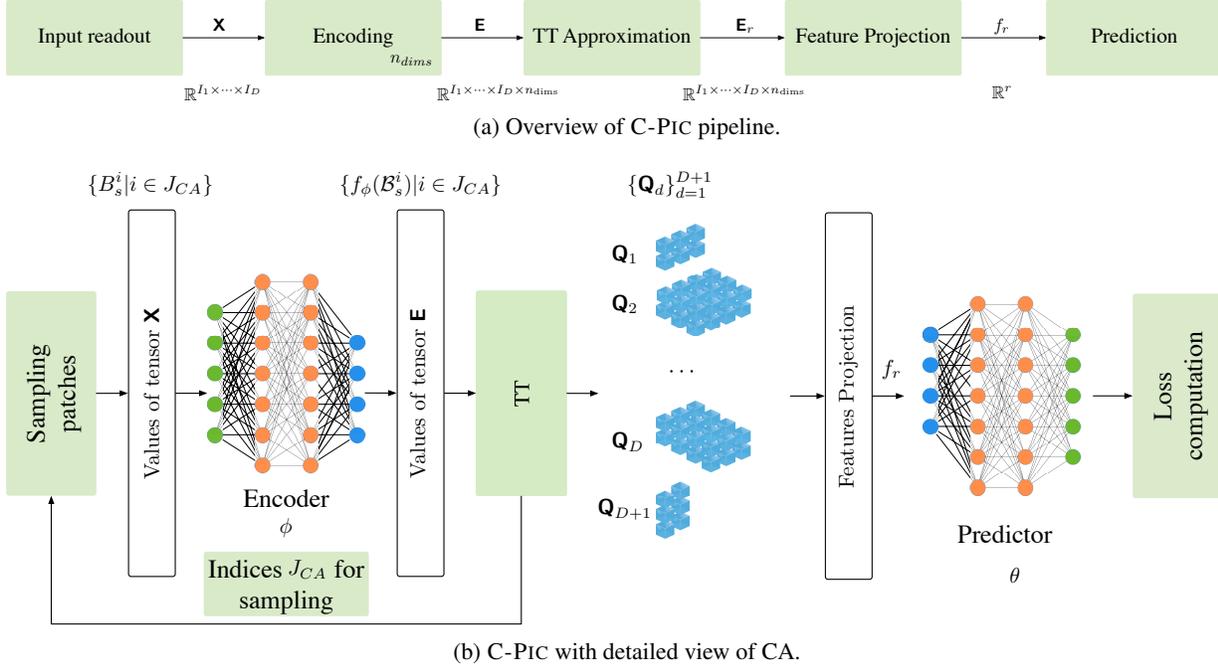

		\centering
		\begin{subfigure}[b]{\linewidth}
			\includegraphics[width=.98\textwidth]{figures/framework1.pdf}
			\caption{Overview of \acro\ pipeline.}
			\vspace{1em}
		\end{subfigure}
		\begin{subfigure}[b]{\linewidth}
			\includegraphics[width=.98\textwidth]{figures/framework2.pdf}
			\caption{\acro \ with detailed view of CA.}
		\end{subfigure}
		\caption{General model architecture (a), and detailed view, c.f.~\cref{alg:ca_tt} (b). The input tensor $\tX$ is treated as if it were partially observed. The indices $J_{CA}$ obtained via cross-approximation define a set of locations $i$ in $\tX$, and the local encoder function processes a voxel-cube  $\mathcal{B}^i_s$ 
		around each of them and outputs feature vectors for the corresponding locations $i$ in $\tE$. These values are used to construct the TT approximation of $\tE$.}
		\label{fig:model}
	\end{figure*}

	\acroshort\ consists of four main building blocks: \emph{(i)} an encoder that can be seen as a learned, non-linear dimensionality reduction; \emph{(ii)} the TT decomposition, followed by \emph{(iii)} feature projection; and \emph{(iv)} a conventional, learned prediction function. See \cref{fig:model}.
	In the first block, a learned mapping transforms the input tensor $\displaystyle \tX$ to a latent encoding $\displaystyle \tE$. The low (tensor) rank of that encoding is imposed by subsequent TT decomposition. This mapping is implemented as a 3D convolutional network (but another differentiable feed-forward operator could also be used). 
	As a result, we obtain for each location in the input tensor $\displaystyle \tX$  a vector in the non-linear encoding $\displaystyle \tE$, i.e., the two tensors have the same shape except for an extra channel dimension in $\displaystyle \tE$. 
	
	In the second block, which has not got any learnable parameters, the encoding $\displaystyle \tE$ is decomposed into a set of TT-cores $\{\tQ_d\}_{d=1}^{D+1}$ with predefined, low TT-ranks, all bounded by a hyper-parameter $r$. The rank $r$ constrains the effective capacity of the representation $\displaystyle \tE$ and offers a trade-off between expressiveness and memory constraints. Crucially, to build the TT decomposition one need not store the full tensors $\displaystyle \tX$ and $\displaystyle \tE$ in memory, rather it is sufficient to observe them at specific locations as described in \cref{sec:efficient_learning}. This makes it possible to sidestep memory limits, but poses the challenge of propagating gradients through the selection of discrete locations. 
	
	In the final two blocks, the obtained TT-cores are used as a basis for the prediction. Since the TT decomposition is not unique, they are first projected onto a canonical basis to obtain an invariant feature vector (see \cref{subsec:feature_projection}), which then serves as input for the final prediction step, in our implementation a multi-layer perceptron (MLP). 
	%
	
	\begin{figure}[tb]
		\centering
		\includegraphics[width=\linewidth,trim=0 0 0 450, clip]{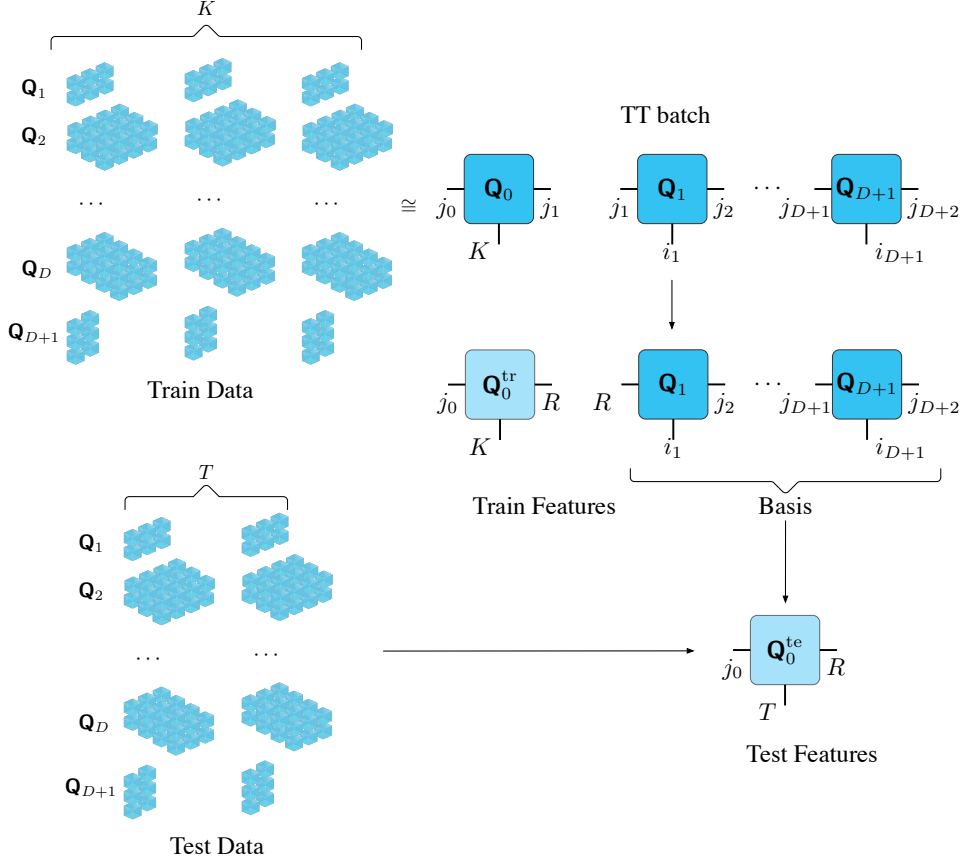}
		\caption{Feature projection. We follow the notation from \cite{cichocki2016tensor}: each blue box represents a TT-core (3$^\text{rd}$-order tensor). The leading and trailing dimensions satisfy $j_0 = j_{D+2} = 1$. We extract invariant features for the $K$ training instances by stacking and rank-truncating them (like PCA for 2D matrices). This yields $K$ feature vectors (core $\displaystyle \tC_0^{tr}$) and an orthogonal basis (cores $\displaystyle \tC_{1},\hdots, \tC_{D+1}$).}
		\label{fig:projection}
	\end{figure}
	
	
	
	\vspace{0.5em}
	\subsection{Differentiable Cross-approximation}
	\label{sec:efficient_learning}
	
	If the tensors $\displaystyle \tX$ and $\displaystyle \tE$ have high resolution, storing them in memory quickly becomes intractable. 
	Therefore, we propose to utilise an efficient \emph{approximate} tensor learning algorithm termed cross-approximation (CA)~\cite{oseledets2010tt}.
	The principle of CA is to reduce memory consumption by only considering selected entries of the tensor $\displaystyle \tX$, at carefully chosen locations.

Originally, CA was conceived as a matrix sampling method~\cite{tyrtyshnikov2000incomplete,bebendorf2000approximation} that uses the so-called \emph{pseudo-skeleton decomposition}~\cite{goreinov1997theory} to approximately reconstruct a matrix $\mU$ while observing only $r$ of its rows and columns. The intersection of these rows with indices $J_1$ and columns with indices $J_2$ define an $(r \times r)$-sized submatrix $\mU(J_1,J_2)$. Finding $J_1,J_2$ that yield the largest $|\text{det}(\mU(J_1,J_2))|$ leads to a rank-$r$ matrix interpolant $\mU(:,J_2)\mU(J_1,J_2)^{-1} \mU(J_1,:)$ with the (up to a constant factor) lowest approximation error w.r.t. the original $\mU$~\cite{goreinov01maxvol}.

	%
	The same idea can be applied in $D\!>\!2$ dimensions as well: a small subset of tensor indices can, under reasonable conditions~\cite{oseledets2010tt}, be used to approximate the tensor $\tE$, which in turn gives rise to an approximate TT decomposition $\{\displaystyle
	\tQ_d\}_{d=1}^{D+1}$ of $\tE$.

	Let $J_{CA}$ be a set of some $N$ locations in the tensor $\tE$ with $D+1$ dimensions, i.e., $J_{CA} = \{(i_1^n, \dots i_{D+1}^n)\}_{n=1}^{N}$. CA alternates between two steps of choosing the indices $J_{CA}$ and building the TT-cores $\{\tQ_d\}_{d=1}^{D+1}$ as follows:
	\begin{enumerate}[leftmargin=15pt,topsep=4pt]
		\setlength{\itemsep}{2pt}
		\setlength{\parskip}{2pt}
		\item \textit{Index selection}: select a set of indices $J_{CA}$ along all tensor dimensions, such that the approximation error is small. The error minimisation is a combinatorial problem and is in practice solved via the greedy \emph{maxvol} heuristic~\cite{goreinov2010find,savostyanov14maxvol}. 
		\item \textit{Cross-interpolation}:  compute TT-cores $\{\tQ_d\}_{d=1}^{D+1}$ based on the entries of $\tE$ evaluated only at indices $J_{CA}$. The cores $\{\tQ_d\}_{d=1}^{D+1}$ are derived from the pseudo-skeleton reconstruction via standard matrix operations, including QR factorization, matrix multiplication, and least-squares inversion. See \cref{app:cross_interpolation} for further details about the CA procedure.
	\end{enumerate}
	
	The value of $\displaystyle \tE$ at a location $i \in J_{CA}$ is obtained by encoding the corresponding local voxel cube $\mathcal{B}^i_s$ from $\displaystyle \tX$, centred at location $i$. In this way, one avoids having to store the full tensors in memory, instead one must only access a set $\{\mathcal{B}^i_s\}$ of $N$ voxel cubes. The fixed, small size $s$ of each cube determines the local context included around each sample and depends on the receptive field of the encoder, see Alg.~\ref{alg:ca_tt}.
	
	When used to approximate a given tensor $\tE$ in a classical way, CA iterates only over the index selection step, then explicitly assembles the TT decomposition of $\tE$ with cross-interpolation. Our work is the first to employ CA within a larger, trainable neural architecture. This means that, during training, the source $\tE$ changes in response to the evolving encoder weights. Consequently, the set of indices $J_{CA}$ must also be updated throughout the learning process. While cross-interpolation consists of differentiable algebraic operations, index selection is a discrete function that poses a problem when training the pipeline end-to-end. To overcome this issue, we propose a scheme that alternates between iterative index selection and gradient descent. More specifically, we cherry-pick the tensor elements and the associated gradients as follows: 
	First, we select and fix a set of indices $J_{CA}$ and, using those, perform back-propagation through the cross-interpolation procedure to update the network weights.
	Then, to catch up with the changed encoder parameters and associated representation $\tE$, we pick a new set of indices $J_{CA}$, switch back to back-propagation at those new locations, and so on. It is easy to see that this procedure converges since, for a given input, the index selection no longer changes once the encoding has converged.

	\textbf{Complexity of CA.} It can be shown~\cite{oseledets2010tt} that an index set $J_{CA}$ containing $N (r) = O(D r^2 \max_d[I_d])$ entries from $\tE$ is sufficient to interpolate $D$ cores, and respectively
	$O(D r^2 \max_d[I_d]s)$ entries from $\tX$. 
	Each of the TT-cores $\{\tQ_d\}_{d=1}^{D+1}$ contains $r^2 I_d$ elements, thus storing them does not change the memory complexity. 
	The time complexity of the cross approximation algorithm (without the cost of sampling the tensor elements) is $O(D r^3 \max_d[I_d])$ \cite{oseledets2010tt}.
		
	\hspace{-5.5mm}
		\begin{algorithm}[!t]
			\centering
			\caption{\textsc{Differentiable CA for TT}}
			\begin{algorithmic}[1]
				\vspace*{.18cm}
				\STATEx \hspace*{-.6cm} $\sI = I_1 \times I_2 \times I_3$ -- 3D grid 
				\STATEx \hspace*{-.6cm} $\tX \in \R^{\sI}$ -- input visual data
				\STATEx \hspace*{-.6cm} $\tE \in \R^{\sI \times n_{\text{dims}}}$ -- full-rank output of encoder $f_{\phi}$
				\STATEx \hspace*{-.6cm} $\mathcal{B}_s^i \subset \sI \times n_{\text{dims}}$ -- $s$-neighborhood of $i \in \sI \times n_{\text{dims}}$
				\vspace*{.1cm}
				\REQUIRE Input data $\tX$, local size $s$
				\FOR{$\text{epoch} = 1, \dots, n_{\text{epochs}}$}
				\FOR{$d = 1, 2, 3, 4$}
				\STATE Select CA indices $J_{CA} \subset \sI \times n_{\text{dims}}$ (Sec.~\ref{sec:efficient_learning})
				\ENDFOR
				\FOR{$d = 1, 2, 3, 4$}
				\STATE Get $\tX[j], \ \forall j \in \{ \mathcal{B}_s^i |\  \forall i \in J_{CA} \}$
				\STATE Get $\tE[i] =  f_{\phi}(\mathcal{B}_s^i), \ \ \forall i \in J_{CA}$
				\STATE Compute $\tQ_d$ via \emph{cross-interpolation} (Sec.~\ref{sec:efficient_learning})
				\ENDFOR
				\STATE Project cores $\tQ_1, \displaystyle \tQ_2, \displaystyle \tQ_3, \displaystyle \tQ_4$ into lower-dimensional features $f_r$ (Sec.~\ref{subsec:feature_projection})
				\STATE Compute loss $l$ of $f_r$
				\STATE Update cores via $\text{backprop}(l)$
				\ENDFOR
				\STATEx {\bfseries Output:} $
				\tQ_1, \displaystyle
				\tQ_2, \displaystyle
				\tQ_3, \displaystyle
				\tQ_4$
			\end{algorithmic}
			\label{alg:ca_tt}
		\end{algorithm}
	
	
	\subsection{Feature projection} \label{subsec:feature_projection}
	
	The TT decomposition is, by construction, not unique.%
	\footnote{E.g., one can create an equivalent TT with different weights by right-multiplying all slices of some core with any non-singular matrix $\displaystyle \mR$ and left-multiplying all slices of the subsequent core with its inverse  $\displaystyle \mR^{-1}$.} %
	%
	To address this issue, our pipeline includes a PCA-like step that projects the TT-cores into a canonical feature space of rank $r$ as follows.
	Given multiple training instances $k=1\hdots K$, we view their TT decompositions $\{\{\displaystyle\tQ^k_d\}_{d=1}^{D+1}\}_{k=1}^K$ as a set of $K$ vectors that forms a basis. We concatenate these vectors in the TT format along a new, leading dimension to form a $(D+2)$-dimensional TT tensor $\tC$ representing that basis, i.e. $\tC[k, ...] \approx \tQ_k$ (the concatenation is done in the TT compressed domain). The first core $\tC_0$ of $\tC$ has shape $1 \times K \times j_1$, i.e. it indexes the training instances along its spatial dimension. Next, we orthogonalise $\tC$ with respect to $\tC_0$ and rank-truncate the resulting core into a $K \times r$  feature matrix $\tC^{\text{tr}}_0$. The trailing cores $\{ \tC_d\}_{d=1}^{D+1}$ now form an orthogonal basis, while matrix $\tC^{\text{tr}}_0$ contains one $r$-dimensional feature vector $\displaystyle\vf_k$ for each input $\displaystyle \tX_k$ that is now invariant to the choice of coefficients in the TT representation $\tQ_k$. The whole procedure is an extension of standard PCA matrix projection to the case where basis elements are TT tensors; see also \cref{fig:projection}. For inference, we similarly concatenate input instances into a new tensor, which we then project onto the learned basis to obtain their corresponding $r$-sized feature vectors. We refer to the supplementary material for further details.
	
	\subsection{Technical Details}
	
	\noindent \textbf{Tensorisation.} An important technical detail along the way is the shape of the embedding $\tE$ that affects the memory complexity. 
	In principle, one can directly apply TT decomposition to the tensor $\tE$ sampling and storing $O(D r^2 \max_d[I_d]s)$ entries of it.
	However, if the tensor has high spatial resolution, i.e., $\max_d[I_d]$ is large, one can reach better memory complexity by employing the so-called Quantised\footnote{The name does not imply quantisation of real-valued tensor entries.} Tensor Train (QTT) decomposition~\cite{khoromskij2011dlog,osel-2d2d-2010}. 
	
	The idea of QTT is to build a TT decomposition for the tensor after reshaping it to a higher dimensional one.
	Particularly, if all $\{\log_2 I_d\}_{d=1}^D$ are natural numbers,  a $D$-dimensional tensor $\displaystyle \tE$ with sizes $\{I_d\}_{d=1}^D$ can be reshaped into a $D'$-dimensional tensor $\displaystyle \tilde\tE$ with $D'= \sum_d \log_2 I_d$ and sizes $\{I_d=2\}_{d=1}^{D'}$. As the result, QTT decomposition of $\tE$ requires a storage cost of $\mathcal{O}(r^2 D \max_d [\log_2 I_d]s)$, as opposed to  the initial $\mathcal{O}(r^2 D \max_d [I_d]s)$. Intuitively, the QTT scheme exploits the similarity between adjacent voxels in the uncompressed tensor $\displaystyle \tE$ and is related to the wavelet transform; see, e.g., \cite{OT:11}.
%
	%
Note that the reshaping is only done locally and implicitly within the QTT routine, by a function that maps index tuples from $\displaystyle \tE$ to $\displaystyle \tilde\tE$ and vice versa.

	QTT is the most sample-efficient scheme for tensors with large $\max_d [I_d]$  and we exploit it in \cref{sec:experiments} to handle resolutions that are intractable with standard deep learning models. 
Still, our scheme is flexible. If the number of samples is not a concern, one can use the conventional TT representation without reshaping during CA. In principle, it is also possible to use our scheme with the exact TT-SVD algorithm instead of the approximate CA to find the decomposition, if the inputs are small enough to fit them into memory. 

	\noindent \textbf{Feature projection and batching.} For PCA, the number of samples $K$ must be at least as large as the feature dimension $r$. Consequently, the batch size during learning must be at least $r$ samples per mini-batch. Note that the common basis would in principle have to be computed over all training samples. In practice we cache the basis in each mini-batch and append it to the cores of the next mini-batch to converge towards a stable, common basis at the end of an epoch.
	
	\noindent \textbf{Numerical issues.} The basis computation is implemented with the \emph{cvxpylayer} method~\cite{cvxpylayers2019}, which we found to have better stability than other algebraic schemes during the backward pass through our differentiable CA.
	Due to the memory-efficiency of \acro\ we are also able to train it with \emph{float64} precision to further improve numerical stability. We do this in all experiments (baselines had to be trained with \emph{float32} to stay within memory limits).

    \noindent \textbf{Index batching.}
	A subtle technical detail is that performing the backward pass simultaneously for all indices selected by CA still requires significant memory, especially for large inputs that require more CA samples.
	To further reduce memory consumption, one can switch to batch-wise processing of the CA indices, such that only the gradients for one batch must be held in memory.
	 However, the price to pay is an increase in runtime, proportional to the number of batches, as one has to run the cross-interpolation step more often.
We have implemented the index batching trick and have empirically verified convergence for tensors up to size $8192^3$.
Still, we recommend to use index batching only when necessary, as it greatly slows down the training (and even for scans of size $512^3$ the complete backward pass fits into the memory of a modern GPU).

\section{Experiments}
\label{sec:experiments}
To illustrate the effectiveness of \acroshort, we apply it to two different prediction tasks where health indicators shall be regressed from medical 3D scans.
The tasks were selected because of their global, holistic nature, i.e., in both cases one should assess the state of an entire organ and the future progression of the condition, for which it makes sense to process the entire scan, rather than cut it into smaller tiles.

\subsection{Datasets}

\begin{figure}[t]
		\centering
		\begin{subfigure}[b]{0.45\columnwidth}
			\includegraphics[width=\linewidth]{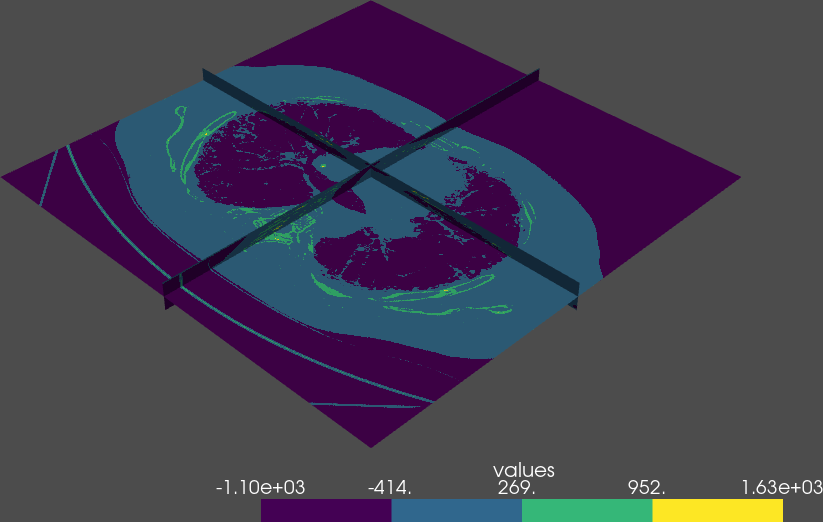}
		\end{subfigure}	
		\begin{subfigure}[b]{0.45\columnwidth}
			\includegraphics[width=\linewidth]{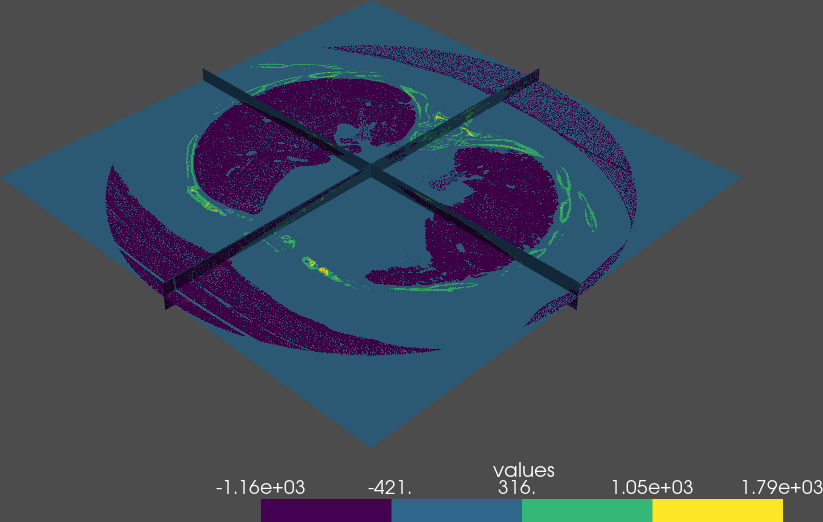}
		\end{subfigure}	
		\begin{subfigure}[b]{0.45\columnwidth}
			\includegraphics[width=\linewidth]{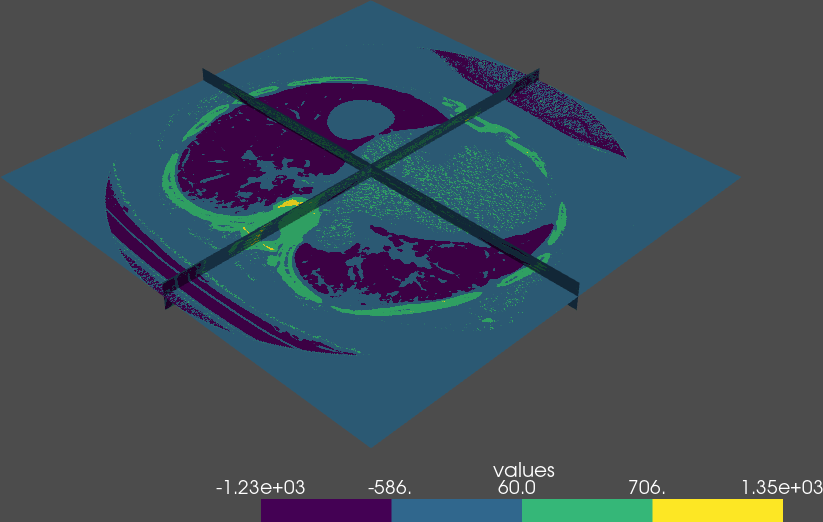}
		\end{subfigure}	
		\begin{subfigure}[b]{0.45\columnwidth}
			\includegraphics[width=\linewidth]{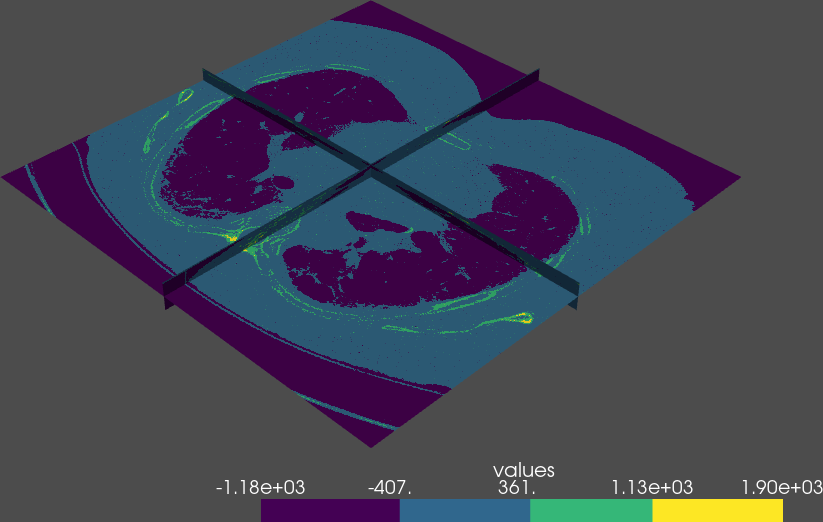}
		\end{subfigure}	
		\caption{Examples from OSIC, resolution $32\times512\times512$.}
		\label{fig:osic_data2}
	\end{figure}

\noindent \textbf{OSIC Pulmonary Fibrosis Progression} is a dataset of CT scans of patients' lungs, originally released for a Kaggle competition~\cite{OSICweb}. Example scans are shown in \cref{fig:osic_data2}.
Pulmonary fibrosis causes a progressive decline of the pulmonary capacity, and the goal of the challenge is to predict that decline from a scan taken at time $t\!=\!0$.
Lung capacity is quantified by forced vital capacity (FVC, the volume of exhaled air exhaled). For the patients in the dataset it has been repeatedly measured over 1-2 years after the scan by means of a spirometer. FVC as a function of time (in weeks) is the regression target.
Overall, there are 176 patients and 1549 individual ground truth FVC values.
%
As error metric, the creators of the challenge proposed the modified Laplace Log Likelihood (mLLL), defined as $\text{mLLL} =-\sqrt{2}\Delta/\sigma-\ln(\sqrt 2\sigma)$; with $\sigma$ the standard deviation of the predicted FVC, truncated at 70 FVC units, and $\Delta$ the absolute error of the predicted FVC, truncated at 1000 FVC units.
For training we use quantile regression. From predicted $\{0.2, 0.5, 0.8\}$ quantiles we compute both predicted FVC and standard deviation.  

\begin{figure}[t]
		\centering
		\begin{subfigure}[b]{0.45\columnwidth}
			\includegraphics[width=\linewidth]{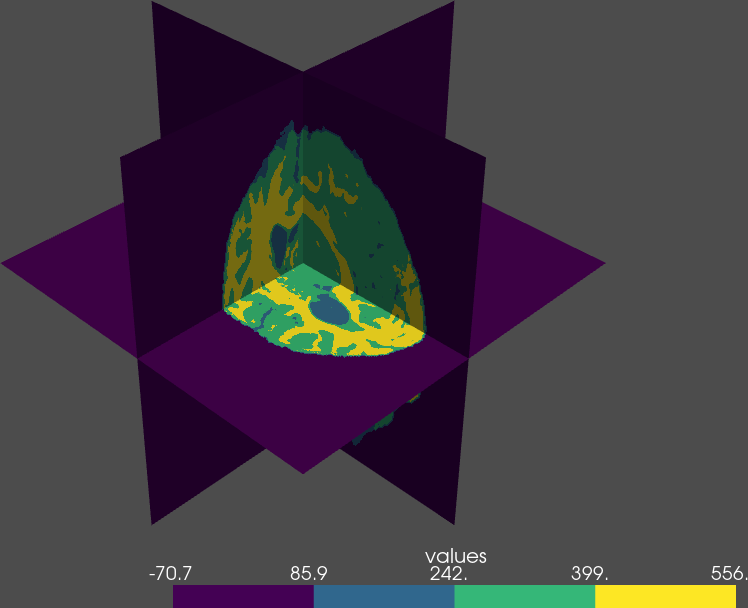}
		\end{subfigure}
		\begin{subfigure}[b]{0.45\columnwidth}
			\includegraphics[width=\linewidth]{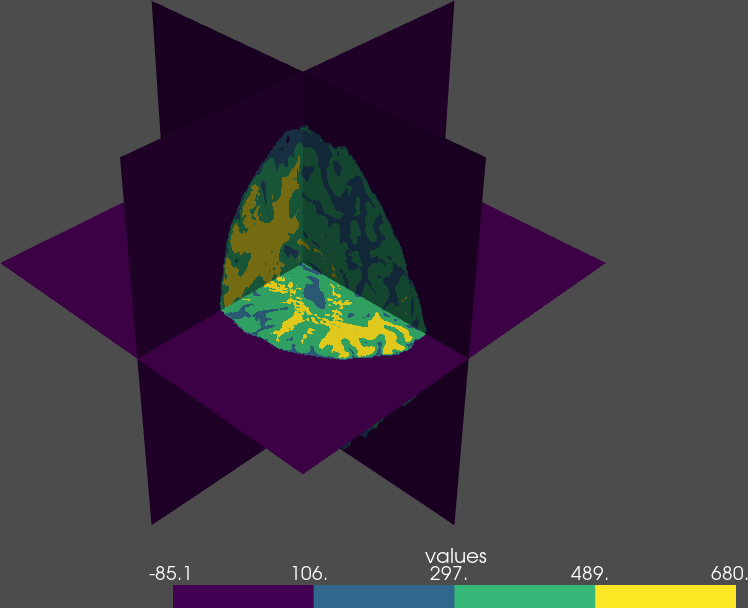}
		\end{subfigure}	
		\begin{subfigure}[b]{0.45\columnwidth}
			\includegraphics[width=\linewidth]{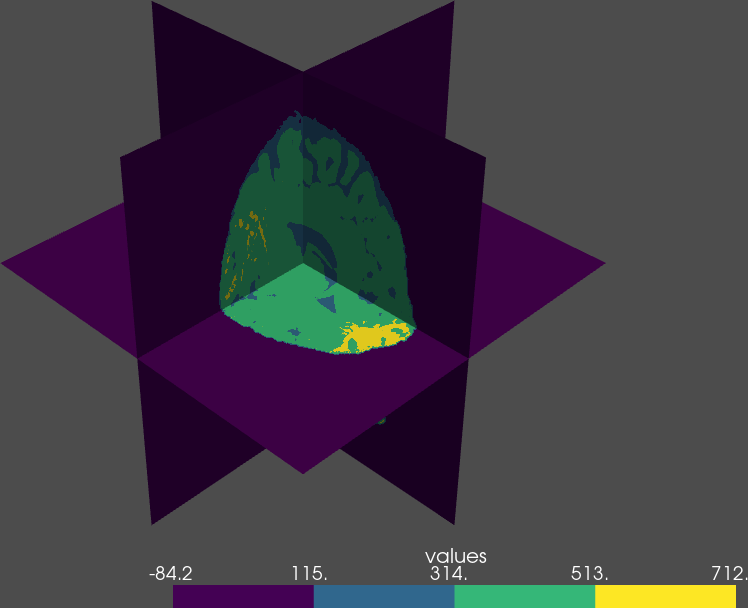}
		\end{subfigure}	
		\begin{subfigure}[b]{0.45\columnwidth}
			\includegraphics[width=\linewidth]{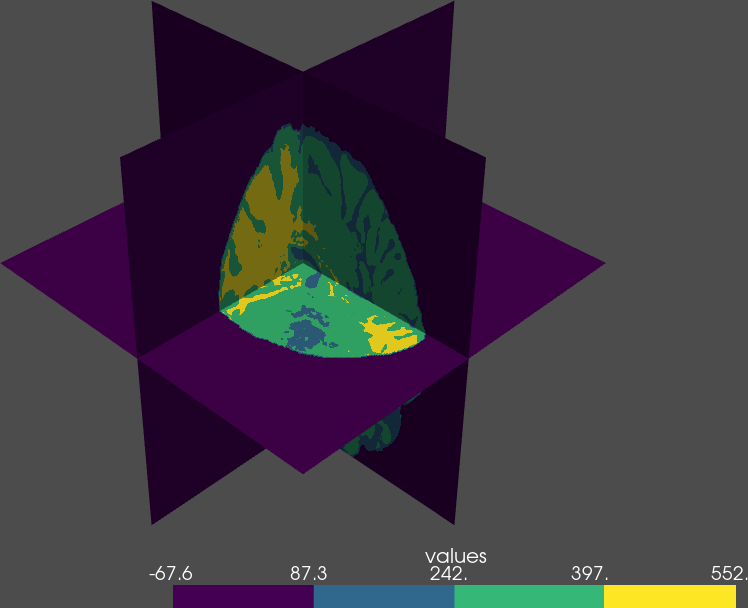}
		\end{subfigure}	
		\caption{Examples from BraTS, resolution $256\times256\times256$.}
		\label{fig:brats_data_3d}
	\end{figure}

\noindent \textbf{MICCAI 2020 BraTS} is a dataset of MRI scans~\cite{BRATSweb} showing brains with a specific type of tumor~\cite{menze2014multimodal,bakas2017advancing,bakas2018identifying}. Examples are shown in \cref{fig:brats_data_3d}.
The target value that should be predicted from a scan is the patient's survival time (in days) after the scan was taken.
The participants of the study are divided into two groups, where the first group underwent a specific type of treatment (gross total resection surgery), whereas the second group was not.
In total there are for 235 patients. We discard grouping and treat all scans as one single dataset for survival prediction.
As error metric, we use the RMSE of the predicted survival time.
During training we normalise the survival time to 5 years.

\noindent \textbf{Synthetic upsampling.} The goal of our work is efficient, compressed representation learning that is able to handle large, high-resolution data. However, there do not seem to be any high-resolution benchmark datasets of sufficient size (although modern scanners can capture up to at least 
$1024\times1024\times128$ voxels).
Therefore, we also synthetically upsample the two datasets to 2$\times$ higher resolution along each dimension with 3\textsuperscript{rd}-order spline interpolation, to obtain 8$\times$ higher voxel count. Clearly, this step does not add any information to the lower-resolution originals, so we do not expect better performance, still the upsampled version gives us an opportunity to verify that our approach can indeed handle such large volumes. In fact, it does so without any loss of accuracy, which supports the hypothesis that the data has low rank and can therefore be compressed without information loss.

\noindent \textbf{Implementation Details.} The detailed layer structure of the CNN encoder and the MLP for prediction are given in~\cref{sec:appendix_nn}. All models are trained with RAdam~\cite{liu2019radam}, with base learning rate $10^{-3}$. All quantitative results are averages over five-fold cross-validation.

\subsection{Results}
	
We first apply \acro\ to the data in its original form (without upsampling), and compare it to a 3D version of ResNet-34~\cite{he2016deep} as a baseline. \acro\ is trained with batch size 20, for the baseline we had to reduce the batch size to 2 to fit the training into memory.
To show that the CNN encoder is indeed needed before the low-rank constraint can be imposed, we also run our pipeline without the encoder. I.e., the raw input tensor $\tX$ is fed into TT decomposition, projected to a canonical feature vector, and fed into the regression MLP.

Quantitative results for the \textbf{OSIC} dataset are shown in \cref{tab:osic}. They show that \acro, with rank $r=10$ and channel depth $n_{\text{dim}}=16$, not only needs a lot less memory, but in fact predicts FVC significantly better than the ResNet baseline.
The performance gain provides evidence that the low-rank assumption underlying our method is justified, at least for the medical scan data we have used: if the intrinsic rank of the data were higher, there would have to be at least some performance loss; whereas if the assumption is valid, it can even act as a regulariser for the learning process.
The TT+MLP baseline, on the contrary, performs a bit worse than ResNet and significantly worse than \acro, i.e., there appears to be a clear benefit in non-linearly transforming the scans to a "TT-friendly" representation with the convolutional encoder, and consequently in the associated end-to-end learning framework.


    \begin{table*}[t!]
		\centering
		\caption{
			OSIC Pulmonary Fibrosis Progression results. \acro\ outperforms the baselines, and can also handle 8$\times$ larger scan volumes, contrary to a 3D ResNet (marked as N/A in the table).}  
		\begin{tabular}{llc|cccccc}
			\toprule
			& resolution & mLLL$\uparrow$ & training time & prediction time & fw/bw memory & \# params\\
			\midrule
			ResNet 34 & $32\times\,\,\,512\times\,\,\,512$ & -6.86 & $\,\,\,$4650 s. / epoch & $\,\,\,$0.2 s. / sample & $\,\,\,$7.0 Gb & 67M \\
			ResNet 34 & $64\times1024\times1024$ & N/A & N/A & N/A & 57.9 Gb & 67M \\
			TT + MLP & $32\times\,\,\,512\times\,\,\,512$ & -6.91 & 27534 s. / epoch & 14.9 s. / sample & $\,\,\,$1.0 Gb & 64K \\
			\acroshort & $32\times\,\,\,512\times\,\,\,512$ & \bf{-6.73} & 51480 s. / epoch & 25.2 s. / sample & $\,\,\,$3.5 Gb & 87K \\
			\acroshort & $64\times1024\times1024$ & \bf{-6.73} & 62478 s. / epoch & 46.1 s. / sample & $\,\,\,$4.2 Gb & 87K \\
			\bottomrule
			\vspace{0.5em}
		\end{tabular}
		\label{tab:osic}
	\end{table*}
    
    \begin{table*}[t!]
		\centering
		\caption{
			MICCAI 2020 BraTS  results. \acro\ outperforms the baseline in terms of RMSE of the predicted survival time. Additionally, the table also shows \acro\ results with different channel depth of the encoding $\tE$. Reducing the channel depth too far hurts performance, even with the same tensor rank $r\!=\!10$.}
		\begin{tabular}{lcc|cccccc}
			\toprule
			& $\!\!\!\!\!\!\!\!$resolution & RMSE$\downarrow$ & training time & prediction time &  fw/bw memory & \# params\\
			\midrule
			ResNet 34 & $256^3$ & 48.7 days & $\,\,\,\,\,\,$519 s. / epoch & $\,\,\,$0.3 s. / sample & 14.0 Gb & 67M \\
			TT + MLP & $256^3$ & 83.9 days &  $\,\,\,\,\,\,$646 s. / epoch &  $\,\,\,$2.9 s. / sample &  $\,\;$4.2 Gb & $\,\,\,$3K \\
			\acroshort \ $n_{dim} = 32, r = 10$ \ & $256^3$ &\textbf{48.2} days & $\,\,\,$3300 s. / epoch & 13.4 s. / sample & $\,\;$8.9 Gb & 37K \\
			\acroshort \ $n_{dim} = 16, r = 10$ & $256^3$ & 49.1 days & $\,\,\,$2979 s. / epoch & 12.8 s. / sample & $\,\,\,$8.7 Gb & 27K \\
			\acroshort \ $n_{dim} = \,\,\,8, r = 10$ & $256^3$ & 51.1 days & $\,\,\,$2883 s. / epoch & 12.1 s. / sample & $\,\,\,$7.9 Gb & 21K \\
			\acroshort \ $n_{dim} = \,\,\,8, r = 10$ & $512^3$ &  51.2 days & 16560 s. / epoch & 79.0 s. / sample & 45.5 Gb & 21K \\
			\acroshort \ $n_{dim} = \,\,\,8, r = 12$ & $256^3$ &  51.1 days & $\,\,\,$5520 s. / epoch &  27.9 s. / sample & 13.4 Gb & 21K \\
			\acroshort \ $n_{dim} = \,\,\,8, r = 15$ & $256^3$ &  51.1 days & $\,\,\,$7140  s. / epoch & 35.8 s. / sample & 18.6 Gb & 22K \\
			\bottomrule
		\end{tabular}
		
		\label{tab:brats}
	\end{table*}

As a next step, we perform the same experiment with the up-scaled scans to see how our method scales to larger volumes. At the increased size of $64\times1024\times1024$ voxels the ResNet baseline can no longer be trained, as even a last-generation GPU with 24GB of on-board RAM runs out of memory already with batch size 1.
On the contrary, \acro\ reaches the same performance as for the smaller scans (as mentioned earlier, no improvement is expected, since the synthetic up-sampling, in contrast to actual high-resolution scanning, does not add information).
The table also shows that the huge memory savings of \acro\ of course come at the price of longer training time because of the added complexity to back-propagate through the TT bottleneck and CA algorithm.
The difference is partly due to our implementation being not nearly as optimised as standard back-propagation code; but we cannot at this point quantify the speed-up achievable with a careful implementation.
Note, however, the training time grows sub-linearly with the resolution, due to the favourable scaling properties of CA.

Results for \textbf{BraTS} are shown in \cref{tab:brats}. For the bigger scan volume and more complex image content of the brain scans, we keep the rank $r\!=\!10$, but increase the channel depth of the encoding to $n_{\text{dim}}
\!=\!32$ as a default.
Again, \acro\ matches the performance of ResNet baseline, with greatly reduced memory consumption. In fact, it even reaches a slightly lower RMSE, but in this case the margin is small and we do not claim to outperform the baseline.
%
%
Additionally, the table also shows the impact of different channel depths in the encoder. Too few channels negatively affect the prediction, whereas too many significantly increase the runtime.
We emphasise that, while adding channels in the latent space increases the representation power of the encoding $\tE$, it only adds a tiny number of weights (for the corresponding convolution kernels).
The added channels can be interpreted as additional dimensions of the encoded data manifold, which make it easier to "flatten". They do not relax the low-rank constraint: independent of the number of channels in its last dimension, the tensor $\tE$ is decomposed into cores $\{\tQ_d\}$ with the same tensor rank $r\!=\!10$. 

We also test the influence of the tensor rank $r$ on performance, with fixed, low $n_{\text{dim}}\!=\!8$. For ranks $r \in \{10, 12, 15\}$ we observe similar performance.
With values $r\!<\!10$ the training tends to become unstable, thus preventing the model from learning. Whereas for $r\!>\!20$ the training went out of memory (but our implementation is not fully optimised, so higher ranks are likely possible).

	

	\section{Conclusion}
	\label{discussion}
	We have developed a neural network architecture that includes the truncated tensor-train decomposition as a low-rank latent representation, and have devised methods to back-propagate through the decomposition.
	Most notably, we have shown how the compressed TT encoding can be learned by cross-approximation, from a sparse set of local samples drawn from suitable locations of the input tensor.
	Thanks to this strategy, there is no need to store the uncompressed input tensor explicitly, which in turn makes it possible to process large, high-dimensional grids that exceed the memory of commodity hardware.
	In experiments on medical CT and MRI scans, we have demonstrated that our \acro\ method matches or even exceeds the performance of a conventional CNN regressor; while using orders of magnitude less memory, thus making it possible to process much larger data volumes, which we expect to increasingly see in the near future as scanning hardware improves.
	While we have, for practical reasons, concentrated on 3D scan data, our method is generic. As long as the requirement of low tensor rank is met (after a non-linear encoding tuned to fit the subsequent decomposition), our method can also be utilised with tensorial data of dimension \textgreater3.
	
	A limitation of \acro\ is that TT decomposition is not robust against translations and rotations of the input data space, i.e., the inputs are implicitly assumed to be roughly aligned (like medical scans). We do not expect it to work as well for arbitrarily shifted and/or rotated inputs, unless the encoder can compensate for such transformations. One possible solution is to actively favour invariance of the encoding during training, for instance by deep supervision or suitable data augmentation. We leave this for future work.

    In this work we have experimented only with regression tasks. However, the low-rank latent embedding that we learn should be equally applicable in combination with other tasks, like classification or segmentation. We speculate that it may even serve as a basis for a generative model.
    
	{\small
		\bibliographystyle{ieee_fullname}
		\bibliography{refs}
	}
	
    \newpage
	\appendix
	\section{Appendix}
	
	In the following we elaborate a number of technical details that had to be omitted from the main paper for lack of space.
	
	\subsection{Background on Tensor Decompositions}
	\label{app:tensor_decompositions}
	
	Low-rank tensor decompositions originated in chemometrics in the 1960ies~\cite{tucker1966some}, and have since the 1990ies attracted renewed interest in linear algebra and signal processing~\cite{kolda2009tensor}. Contrary to matrices, there is no single, obvious and generally valid definition of a tensor's rank. Multiple useful definitions of tensor rank have been proposed, each of which gives rise to a different decomposition.
	One of the earliest was the \emph{canonical decomposition} \cite{hitchcock-rank-1927,harshman1970foundations,carroll1970analysis}, with the key concept of expressing a tensor as the sum of a finite number of rank-one (separable) tensors.
	%
	The decomposition is memory-efficient, but computing it may be numerically unstable; and it lacks a mechanism to control the trade-off between low rank and higher approximation error \cite{ballester2015tensor,oseledets2011tensor}.
	%
	The Tucker decomposition \cite{tucker1963implications} factors a $D$-dimensional tensor into a smaller $D$-dimensional core tensor and $D$ factor matrices.  
	In contrast to canonical decomposition, in the Tucker model one can find the lowest possible rank for a given (approximation) error budget by truncation.
	On the downside, it is suitable only for low-dimensional tensors (typically, $D\leq 4$), as the number of parameters scales exponentially with the dimensionality $D$.
	
	The more recent TT format~\cite{oseledets2010tt} combines important advantages of the CP and Tucker models: like the former, its storage cost grows sub-exponentially with the dimension $D$, while it is numerically stable like the latter. We note that the recent  \emph{tensor chain} (TC)~\cite{cichocki2016tensor} is another related format that is more symmetric than TT and also has sub-exponential storage cost; however, the difficulty of rounding TC tensors makes them inadequate for the feature projection step required in our model to overcome the ambiguity of the decomposition.
	
	\subsection{Algorithm Details for Cross-Approximation}
	\label{app:cross_interpolation}
	
We use tensor train cross-approximation~\cite{oseledets2010tt,savostyanov14maxvol} to learn an approximate TT decomposition of the $(D+1)$-dimensional latent encoding $\tE$. We do this by collecting sets of samples (tensor elements) iteratively, one dimension at a time, in a sweep-like fashion: we traverse dimensions $1$ through $D$, then $D+1$ through $2$, and iterate. At each sweep $k$, sets of \emph{left indices} $L^k_d$ and \emph{right indices} $R^k_d$ are kept for every dimension $d$, in such a way that the set of selected fibers $\tE[L^k_d, :, R^k_d]$ is a tensor of shape $r_{d-1} \times I_d \times r_d$. By organizing those fibers as row vectors, we obtain a matrix $\mE_d$ of shape $(r_{d-1}\cdot I_d) \times r_d$. We then run \emph{maxvol}~\cite{goreinov2010find} on $\mE_d$ to select an $r_d\times r_d$ submatrix $\tilde{\mE}_d$ containing $r_d$ rows of $\mE_d$, such that $|\text{det}(\tilde{\mE}_d)|$ is as large as possible. The newly selected row indices become $L^k_{d+1}$ and are subsequently used to acquire the required samples for the next dimension $d+1$.
	
Once a full index selection sweep has concluded, we apply the cross interpolation formula: for each dimension $d$, we compute $\mE_d {\tilde{\mE}_d}^{-1}$ and reshape the result into a tensor of shape $r_{d-1} \times I_d \times r_d$ that becomes the $d$-th TT core $\tQ_d$. We cast the matrix inversion product as a quadratic optimization problem and find a differentiable solution of it via the package \emph{cvxpylayers}~\cite{cvxpylayers2019}. 

Forward propagation through \acro\ has complexity
$\mathcal{O}\big(F_sDnr^2 + Dnr^3\big)$, where $s$ is the receptive field around a sample and $F_s$ is the cost of a forward pass with a $n_\text{dim}$-channel tensor of size $s^3\!\times\!n_\text{dim}$.
We do not back-propagate through the index selection step, so the complexity of back-prop for the sampled patches is $\mathcal{O}(B_sDnr^2 )$, with $B_s$ the cost of the backward pass for one patch.

	\subsection{Feature Projection Algorithm}
	\label{app:feature_projection}
	
\begin{figure}[th]
		\centering
        \includegraphics[width=\linewidth]{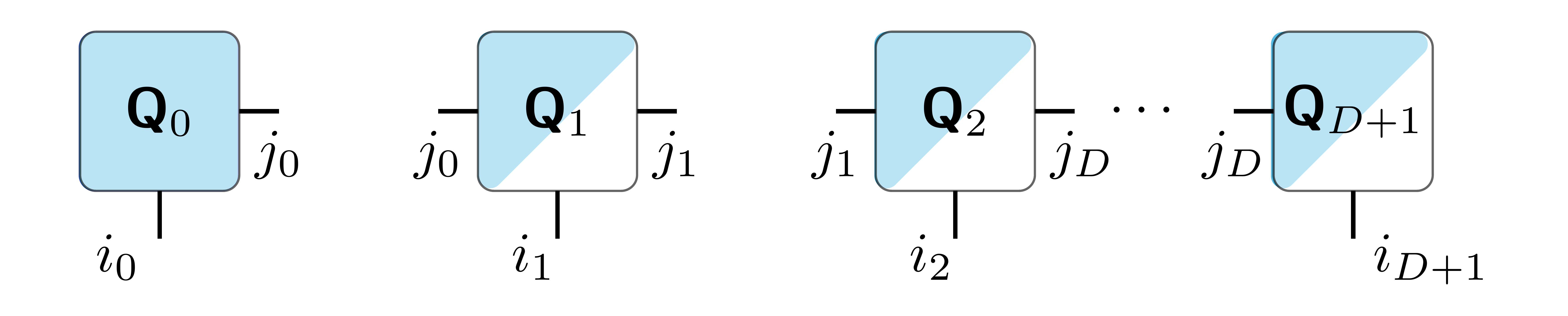}
		\caption{TT decomposition with left-orthogonalised TT-cores $\tQ_d$, $d=1,\dots,D+1$, such that reshaping any $\tQ_d$ into $r_{d-1} \times (I_d r_d)$ yields an orthonormal matrix~\cite{holtz12orthogonalization}. Here, $\tQ_0$ plays the role of $\hat \mU$ from step \ref{step: two}, while the tensor reconstructed from the remaining cores (and reshaped into an $r_0\times (I_1\dots I_{D+1})$ matrix) plays the role of $\hat \mV^T$.}
		\label{fig:orth}
		
	\end{figure}

	Conceptually, our projection step is analogous to the task of computing the principal component analysis (PCA) from a given low-rank matrix factorisation $\displaystyle \mM = \displaystyle \mU  \displaystyle \mV^T$. To accomplish this without explicitly multiplying $\displaystyle \mU$ with $\displaystyle \mV^T$, one can:
    \begin{enumerate}
    \item Compute the QR factorisation of $\displaystyle \mV$, i.e., $\mQ \mR = \mV$
    \item \label{step: two} Set $\hat{\displaystyle \mU} := \displaystyle \mU \displaystyle \mR^T,  \hat{\displaystyle \mV} := \displaystyle \mQ$
        \item Find the principal components of $\hat{\displaystyle \mU}$, which are its $r$ leading left singular vectors.
    \end{enumerate}
    We perform this algorithm in the compressed TT domain, consisting of the following three steps:
    \begin{enumerate}[label=\roman*.]
\item stack the $K$ TT tensors from one training batch along a new, leading dimension to form a new TT tensor of dimension $D+2$ (with dimension $I_0$ the number of samples in the batch);
\item orthogonalise the stack w.r.t.\ that newly formed core (which can be treated as a matrix, since its shape is $1 \times K \times j_1$);
\item find the $r$ leading PCA components of that matrix.
\end{enumerate}
For step ii, instead of a single QR factorisation, our version requires a sequence of QR factorisations and core rotations, see \cref{fig:orth} and~\cite{holtz12orthogonalization}.

	\subsection{Network Architecture Details}
	\label{sec:appendix_nn}
	
	\cref{tab:network_architectures} shows the exact sequence of layers in our (convolutional) encoder, as well as in our (fully connected) regression heads after the TT bottleneck.
	
	 The 3D input scans are mapped to a latent encoding with a convolutional encoder with four layers, \emph{ReLU} activations between layers, and a \emph{sigmoid} after the last layer. All kernels are of size 3$\times$3, resulting in a receptive field of $9^3$ voxels since no padding is used.
	 
	 The regression heads are multi-layer perceptrons, with \emph{ReLU} activations and batch normalisation (for OSIC dataset we found turning off batch normalization leads to better results and faster convergence).
	 For BraTS survival time we employ standard regression, whereas for OSIC Pulmonary Fibrosis progression we use quantile regression to predict uncertainty, moreover we first transform the visual information with two layers without \emph{batchnorm} (left side of center column in \cref{tab:network_architectures}), then fuse it with the week number and decode into the final prediction with three more layers (right side of column).

	\begin{table}[h!]
        \centering
		\caption{Network architectures used in our experiments.}
		\begin{tabular}{l|cc|c}
			\toprule
			Encoder 3D&\multicolumn{2}{c|}{Quantile regressor}& Regressor\\
			\midrule
			conv($3^3, 5$)
			& dense(500) & & dense(100)\\
			ReLU
			& ReLU & & BN\\
			conv($3^3,15$)
			& dense(100) & & ReLU\\
			ReLU
			& ReLU & & dense(20)\\
			conv($3^3,25$) & & dense(100) & BN\\
			ReLU & & BN & ReLU\\
			conv($3^3,n_{\text{dims}}$) & & ReLU & dense(1) \\
			Sigmoid & & dense(20) & \\
			& & BN & \\
			& & ReLU & \\
			& & dense(3)\\
			\bottomrule
		\end{tabular}
		\label{tab:network_architectures}
	\end{table}
	
	\subsection{Selected Visualizations}
	\label{sec:visualisations}
	
\begin{figure}[bt]
		\centering
		\begin{subfigure}[b]{0.45\linewidth}
			\includegraphics[width=\linewidth]{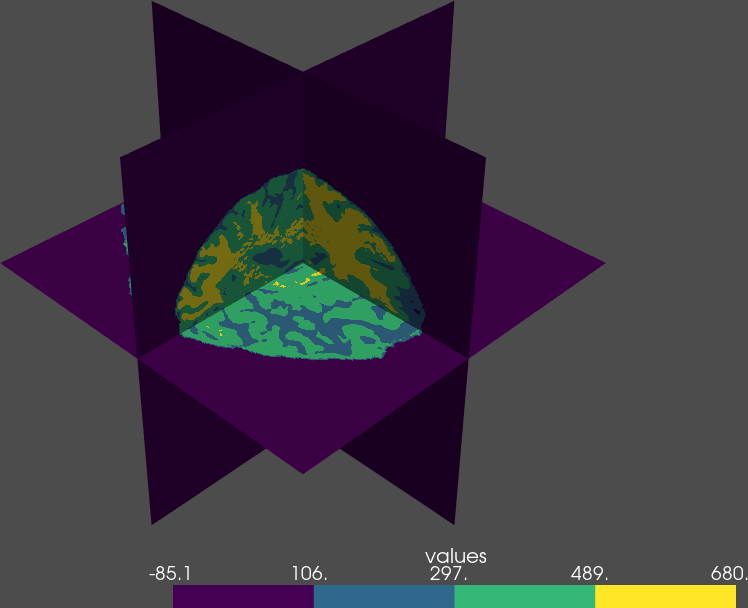}
			\caption{P: 538 days, GT: 616 days}
		\end{subfigure}
		\begin{subfigure}[b]{0.45\linewidth}
			\includegraphics[width=\linewidth]{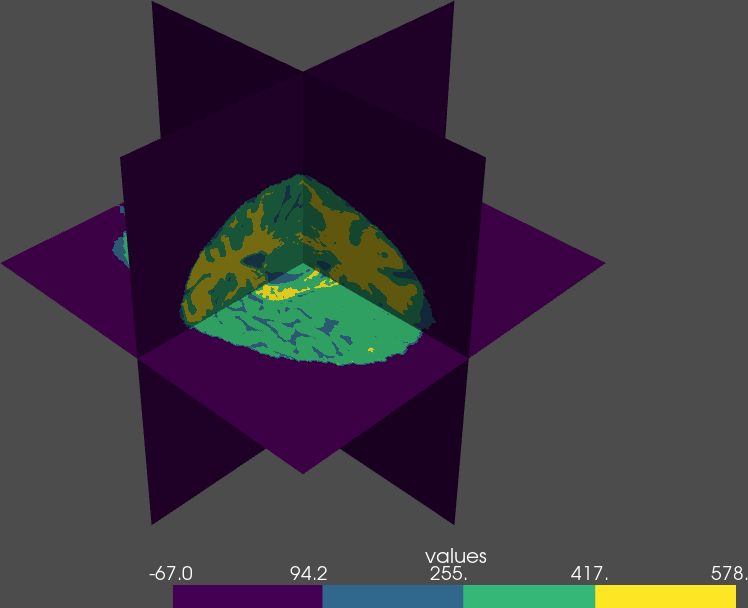}
			\caption{P: 590 days, GT: 515 days}
		\end{subfigure}	
		\begin{subfigure}[b]{0.45\linewidth}
			\includegraphics[width=\linewidth]{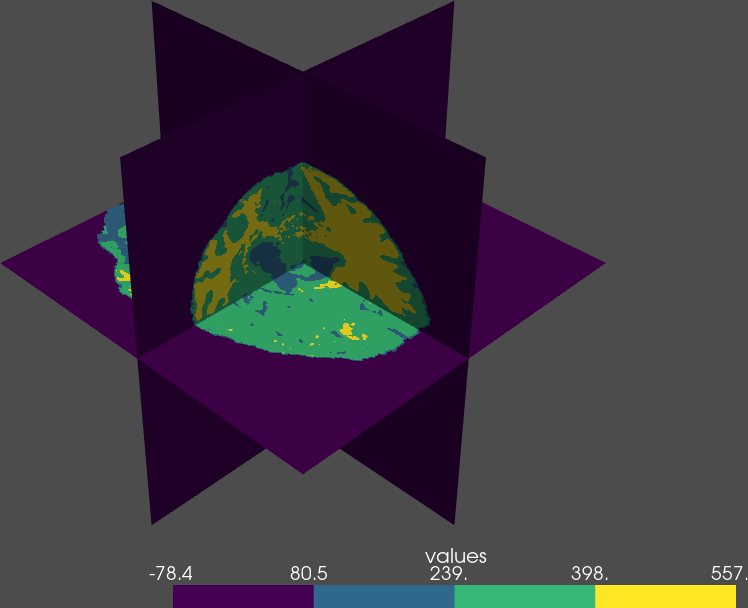}
			\caption{P: 748 days, GT: 698 days}
		\end{subfigure}	
		\begin{subfigure}[b]{0.45\linewidth}
			\includegraphics[width=\linewidth]{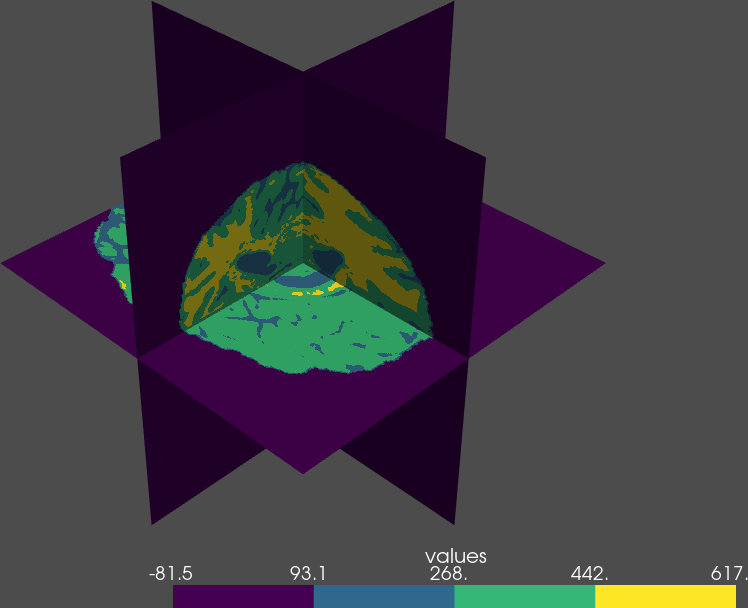}
			\caption{P: 400 days, GT: 359 days}
		\end{subfigure}	
		\begin{subfigure}[b]{0.45\linewidth}
			\includegraphics[width=\linewidth]{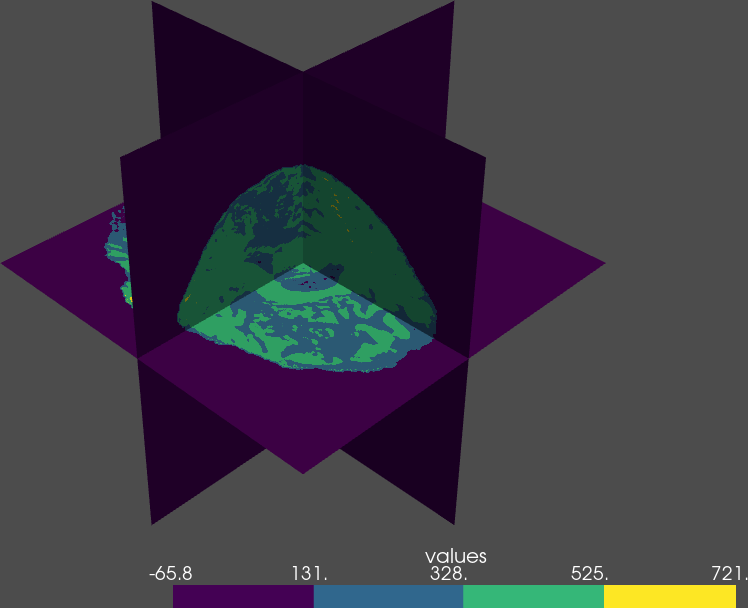}
			\caption{P: 460 days, GT: 439 days}
		\end{subfigure}	
		\begin{subfigure}[b]{0.45\linewidth}
			\includegraphics[width=\linewidth]{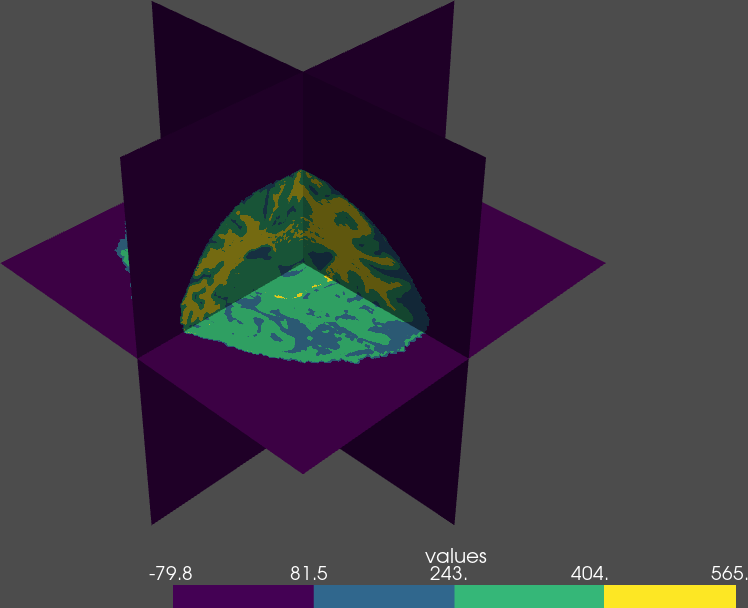}
			\caption{P: 464 days, GT: 486 days}
		\end{subfigure}	
		\caption{Example predictions for BraTS, at resolution $256\times256\times256$.}
		\label{fig:brats_predictions}
	\end{figure}

	\cref{fig:brats_predictions} demonstrates randomly selected predictions of survival time with the trained network on BraTS dataset.
	
\end{document}